# Autonoma: A Hierarchical Multi-Agent Framework for End-to-End Workflow Automation

**Eslam Reda[1], Maged Yasser[1], and Sara El-Metwally[2, *]**

[1] Artificial Intelligence Program, Faculty of Computers and Information, Mansoura University, Mansoura 35516, Egypt.

[2] Computer Science Department, Faculty of Computers and Information, Mansoura University, Mansoura 35516, Egypt.

[*]Address correspondence to this author at the Department of Computer Science, Faculty of Computers and Information, Mansoura University, P.O. Box: 35516, Mansoura, Egypt; Tel/Fax: +2-010-0439-4871;

E-mails: sarah_almetwally4@mans.edu.eg.

## Abstract

The increasing complexity of user demands necessitates automation frameworks that can reliably translate open-ended instructions into robust, multi-step workflows. Current monolithic agent architectures often struggle with the challenges of scalability, error propagation, and maintaining focus across diverse tasks. This paper introduces Autonoma, a structured, hierarchical multi-agent framework designed for end-to-end workflow automation from natural language prompts. Autonoma employs a principled, multi-tiered architecture where a high-level Coordinator validates user intent, a Planner generates structured workflows, and a Supervisor dynamically manages the execution by orchestrating a suite of modular, specialized agents (e.g., for web browsing, coding, file management). This clear separation between orchestration logic and specialized execution ensures robustness through active monitoring and error handling, while enabling extensibility by allowing new capabilities to be integrated as plug-and-play agents without modifying the core engine. Implemented as a fully functional system operating within a secure LAN environment, Autonoma addresses critical data privacy and reliability concerns. The system is further engineered for inclusivity, accepting multi-modal input (text, voice, image, files) and supporting both English and Arabic. Autonoma achieved a 97% task completion rate and a 98% successful agent handoff rate, confirming its operational reliability and efficient collaboration. The system's source code is available on GitHub at https://github.com/eslam-reda-div/Autonoma, and a comprehensive wiki description can be found at: https://deepwiki.com/eslam-reda-div/Autonoma.

.

## Keywords

Multi-Agent Systems; Workflow Automation; Hierarchical Orchestration; Modular Architecture; Task Decomposition; Autonomous Systems; AI Agents.

# 1. Introduction

The evolution of artificial intelligence has been characterized by successive paradigm shifts, each expanding the capabilities and scope of intelligent systems. From rule-based expert systems to machine learning algorithms capable of pattern recognition and prediction, AI has progressively moved toward greater autonomy and sophistication. The latest advancement in this trajectory is the emergence of Agentic AI, a paradigm that fundamentally redefines the relationship between intelligent systems and their operational environments. Agentic AI represents a class of autonomous systems engineered to pursue complex, long-term goals with minimal human supervision, exhibiting capabilities that extend beyond the reactive nature of traditional AI and the pattern-generation focus of generative models [1-4].

The defining characteristics of Agentic AI include goal-oriented autonomy, adaptive decision-making, and the capacity for sustained operation in dynamic, real-world environments. Unlike classical AI systems, which are constrained by predefined rules and narrow task specifications, or generative AI systems, which excel at content synthesis but lack inherent directness, Agentic AI integrates both generative capabilities and strategic goal-pursuit. This synthesis enables agents to manage multi-layered tasks, navigate uncertainty, and optimize their behavior over extended time horizons without requiring continuous human guidance. The transition from passive, instruction-dependent systems to proactive, self-directed agents marks a qualitative leap in AI functionality, with profound implications for both theoretical understanding and practical application. The architectural foundations of Agentic AI are critical to its operational efficacy [5-8].

In the architectural paradigm for constructing artificial intelligence systems, a critical design decision involves selecting between a monolithic agent and an ensemble of micro-specialized agents. Monolithic agents, typically leveraging a single large language model, provide a straightforward solution for general tasks and rapid prototyping. However, for complex, multi-stage workflows requiring high reliability and precision, a micro-specialized architecture demonstrates significant advantages. This approach decomposes a complex objective into discrete subtasks, each handled by a dedicated, fine-tuned agent, thereby enhancing overall system performance through task-specific optimization, improved error containment, and superior scalability, albeit with increased complexity in orchestration and inter-agent communication. Empirical evidence suggests that the micro-specialized model offers a more robust and efficient framework for advanced, production-level AI applications, aligning with modular design principles proven effective in traditional software engineering [9-11].

Modern agentic systems employ sophisticated frameworks designed to decompose complexity, facilitate coordination, and enable scalability. Multi-Agent systems distribute responsibilities among collaborating autonomous entities, allowing for parallel processing of sub-tasks and emergent collective behavior. The integration of external tools and application programming interfaces extends agents' capabilities beyond their inherent computational boundaries, enabling access to real-time data, specialized algorithms, and domain-specific resources [12-14].

This paper introduces Autonoma, a multi-agent AI assistant system that enables users to automate tasks through an intuitive chat interface. Accessible via a web browser or mobile device, Autonoma moves beyond generating insights to directly executing tasks, generating reports, and managing workflows. This is achieved by orchestrating a team of coordinating and specialized working agents within a controlled runtime environment, which creates a dedicated digital workspace for each task. Our evaluation demonstrates the system's robustness and usability. Autonoma achieved a 97% task completion rate and a 98% successful agent handoff rate, confirming its operational reliability and efficient collaboration. Furthermore, the system proved accessible to non-technical users, and its secure, LAN-confined architecture with defense-in-depth measures provides a solid foundation for safeguarding data and workflows. Collectively, these results validate Autonoma as a reliable, user-friendly, and scalable platform for multi-agent orchestration.

This paper makes the following key contributions to the field of AI-driven automation and human-computer interaction:

1. We propose and implement Autonoma, a structured, hierarchical multi-agent framework specifically designed to translate open-ended natural language prompts into executable, multi-step workflows. This architecture provides a principled approach to decomposing complex user intents and managing the execution lifecycle.

2. A hierarchically orchestrated, modular agent architecture for end-to-end workflow automation. This framework introduces a multi-tiered agent model where a Coordinator validates user intent, a Planner generates structured workflows, and a Supervisor dynamically executes these plans by managing a suite of plug-and-play specialized agents (e.g., for web browsing, file management, coding). This separation of concerns, between orchestration logic and specialized execution, is the core of our modular design. It ensures robustness through active monitoring and error handling, while its extensibility allows new capabilities to be integrated simply by adding new specialized agents to the framework, without modifying the core orchestration engine.

3. We detail the implementation of a fully functional system that operates within a secure, local-area network (LAN) environment. This addresses critical concerns regarding data privacy, security, and operational reliability by ensuring that sensitive data and automation tasks do not traverse external cloud services unless explicitly required.

4. The system is engineered to accept user input through multiple data modalities (text, voice, image, files) and supports both English and Arabic text, contributing to the development of more inclusive and accessible AI-powered automation tools.

By combining these elements, Autonoma provides a significant step forward from conversational AI that merely suggests actions to a proactive system that reliably executes them within a user's native digital environment.

The paper is organized as follows: Section 2 summarizes related work. Section 3 presents the Autonoma architecture, and Section 4 describes its modularity design. Section 5 details the workflow process. Section 6 covers the technology stack, security, and extensibility. Section 7 discusses the experimental results, challenges, and future research directions, and Section 8 concludes the paper.

## 2. Related Work

Classical AI systems are typically designed for narrow, well-defined tasks such as image analysis or language translation, operating within a predetermined scope based on supervised learning. Generative AI, while powerful in pattern recognition and content creation, lacks inherent goal-oriented behavior. Agentic AI synthesizes and surpasses these approaches by integrating generative capabilities with goal-directed autonomy, enabling it to manage intricate, multi-layered tasks without requiring continuous instruction. Table 1 provides a comparative summary of the three paradigms, contrasting their approaches to adaptability, learning type, decision-making, environment interaction, and human intervention [15-18].

**Table 1. A comparative analysis of Artificial Intelligence paradigms: Traditional AI, Generative AI, and Agentic AI[*].**

| Features | Traditional AI | Generative AI | Agentic AI |
|---|---|---|---|
| **Purpose** | Task-specific Automation | Content Creation & Pattern Synthesis | Goal-oriented Autonomy |
| **Human intervention** | High (predefined parameters) | Medium (prompt-driven) | Low (autonomous adaptability) |
| **Adaptability** | Limited | Moderate (within trained patterns) | High |
| **Environment Interaction** | Static or limited context | Reactive to prompts | Dynamic and context-aware |
| **Learning Type** | Primarily Supervised | Self-supervised / Generative Models | Reinforcement and Self-supervised |
| **Decision Making** | Data-driven static rules | Pattern-based generation | Autonomous, Contextual Learning |

* Table adapted from [1] with additional added features for generative AI.

The field of artificial intelligence has undergone a paradigm shift with the advent of large language models (LLMs) such as GPT-4 from OpenAI's [19], which demonstrate remarkable proficiency in natural language understanding and generation. However, these models are fundamentally architected as passive, stateless assistants that respond to user prompts; they lack the intrinsic capability to autonomously orchestrate and execute multi-step tasks in dynamic environments. This limitation underscores a critical research gap:

translating model-based reasoning into goal-directed action. Consequently, the next frontier in AI involves the development of general-purpose agentic systems that can bridge this divide, moving beyond conversational response to proactive task completion through deliberate planning, tool usage, and environmental interaction [20-22].

OpenAI's Operator [23] is a web-based autonomous agent that leverages GPT-4o's [24] vision and advanced reasoning to perceive and interact with graphical user interfaces (GUIs) via a dedicated browser, enabling it to perform tasks like form completion and e-commerce without API dependencies. The agent operates within a safeguarded research preview, featuring self-correction mechanisms and explicit human handoff protocols for sensitive actions (e.g., logins, payments) to ensure safety and iterative refinement.

Anthropic's Computer Use [25] represents a significant advancement in agentic AI, enabling the Claude 3.5 Sonnet model [26] to perceive and manipulate graphical user interfaces through a programmatic action loop. The system operates by iteratively capturing screenshots, interpreting visual elements using its vision capabilities, and executing precise mouse and keyboard commands through defined tools (computer, text editor, bash). This creates a closed-loop agent that can autonomously complete multi-step tasks, from form filling to software development, by continuously evaluating outcomes against stated objectives. While demonstrating substantial potential for workflow automation, the technology currently faces limitations in interaction reliability, latency, and tool selection accuracy, necessitating its deployment within safeguarded environments like Docker containers during this experimental phase.

Google's Mariner [27] is an experimental AI browser agent developed by Google DeepMind that leverages Gemini's [28] multimodal capabilities to autonomously perform web-based tasks. The system operates by visually parsing a browser interface, planning a sequence of actions, and executing low-level interactions such as clicking, typing, and scrolling to complete user-defined objectives like ticket purchasing and form filling. While demonstrating proficiency in visual understanding and task decomposition across multiple tabs, the agent currently faces significant limitations in execution speed, reasoning for optimal decision-making, and robustness against common web obstacles and login barriers.

Monica's Manus AI [29] represents a significant advancement in autonomous AI agents, distinguished by its capacity for end-to-end task execution with minimal human intervention. Its technical architecture enables multi-modal data processing (text, images, code), sophisticated integration with external tools and software APIs, and continuous learning from user interactions for personalization. These core capabilities (i.e. autonomous planning, advanced tool use, and adaptive behavior) allow it to function as a general-purpose assistant capable of automating complex workflows across diverse domains such as healthcare, finance, and software development, positioning it at the forefront of applied agentic AI systems. Tables 2 and 3 Summarize all features analysis between Autonoma and different agentic AI systems.

**Table 2.** Feature Comparison of AI Automation Agents: Autonoma, Manus, Computer Use, and Mariner*.

| Feature | Autonoma | ManusAI | Operator | Computer Use | Mariner |
|---|---|---|---|---|---|
| **Agent Type** | Mobile and Browser-based (operates in Linux sandboxs) | Browser-based (operates in Linux sandboxs) | Browser-based | API-based | Browser-based (Chrome extension) |
| **Autonomous Web Browsing** | Yes | Yes | Yes | Yes | Yes |
| **Form filling and data entry** | Yes | Yes | Yes | Yes | Yes |
| **Online shopping and reservations** | Yes | Yes | Yes | Yes | Likely Yes |
| **Multimodal I/O (Text, Images)** | Yes | Yes | Limited | Limited | Yes |
| **Integration with external API's** | Yes | No | No | Yes | Limited (Browser APIs) |
| **Task Complexity** | Multi-step workflows | Presumably Multi-step | Simple to Moderate | Complex, programmable | Research-focused |
| **Handling Dynamic Content** | Robust | Unknown | Can struggle | Highly adaptable | Research Phase |
| **Memory & Context** | Good session memory | Unknown | Basic session memory | Customizable via code | Unknown |
| **Code/Logic Customization** | Yes (Open Source) | No | No | Yes (API-based) | No |
| **Availability** | Open Source (Github) | Beta (Invite-only) | Subscribers | Beta (API-access) | Research Phase |

* Table adapted from [29], with additional added features based on the Autonoma multi-agent system architecture.

**Table 3. Feature Analysis of Autonoma's Multi-Agent Architecture vs. Monolithic Single-Agent Systems.**

| Feature / Aspect | Autonoma | Monolithic Single-Agent Systems |
|---|---|---|
| **Architecture Type** | Multi-agent system, a collection of separate AI agents, each with a specific task and responsibility. | Single large AI agent that handles multiple tasks using various integrated tools. |
| **Task Distribution** | Each agent focuses on one well-defined job, leading to specialization and accuracy. | One agent tries to manage all tasks, leading to complexity and reduced focus. |
| **Training & Improvement** | Each agent can be trained, modified, or enhanced individually without affecting others. | Training the entire system affects all tasks, difficult to fine-tune one capability independently. |
| **Performance Efficiency** | High — agents are optimized for their specific roles, resulting in faster and more precise execution. | Lower — the agent must manage tool selection and context switching within a long reasoning loop. |
| **Scalability** | Easily scalable — new agents can be added for new tasks or replaced independently. | Hard to scale — adding new tools or capabilities increases loop complexity and error risk. |
| **Error Isolation** | If one agent fails, others continue working normally. | A single agent failure can disrupt the entire workflow. |
| **Hallucination Risk** | Lower — focused agents work on defined data and goals, minimizing context drift. | Higher — long reasoning chains and multitasking increase chances of hallucination and wrong assumptions. |
| **Maintenance** | Modular — easy to update, debug, and monitor individual agents. | Monolithic — hard to diagnose and fix issues inside a complex single-agent loop. |
| **Adaptability** | Each agent can use a different model, dataset, or method suited to its task. | Single model must generalize across all tasks, often leading to weaker performance in some. |
| **Use Case Example** | A coordinator, planner, supervisor, and specialized research agent working autonomously in a coordinated, multi-agent system. | One AI assistant trying to research, analyze, and design at once. |

## 3. Autonoma Architecture

Autonoma is a multi-agent AI assistant system that enables users to automate tasks through an intuitive chat interface. Accessible via a web page or mobile device, it goes beyond providing insights by directly executing tasks, generating reports, and managing workflows. The system transforms user prompts into action by orchestrating a team of specialized agents (see Figure 1), each dedicated to a specific function like web browsing, file management, or desktop automation. Built on a cutting-edge technology stack including *LangChain* and *LangGraph*, Autonoma operates within a secure, LAN-only environment, creating a controlled digital workspace for each task and ensuring efficient and secure automation. The detailed implementation of each agent in Autonoma will be discussed in the following subsystems. The specifications for these agents are summarized in Table 4. This table outlines each agent's underlying model and primary responsibility. The framework is composed of specialized agents that work in concert to enable end-to-end task automation. The *Coordinator* receives and validates user intent, while the *Planner* transforms these requirements into structured workflows. The *Supervisor* then orchestrates execution by routing tasks to the appropriate worker agents: the *Researcher* for external data collection, the *Coder* for scripting, the *Browser* for web automation, the *File Manager* for local file operations, and the *Computer* for desktop control. Finally, the *Reporter* aggregates the results into a structured summary. This modular design ensures scalability, robustness, and transparency, allowing complex tasks to be completed through seamless agent collaboration.

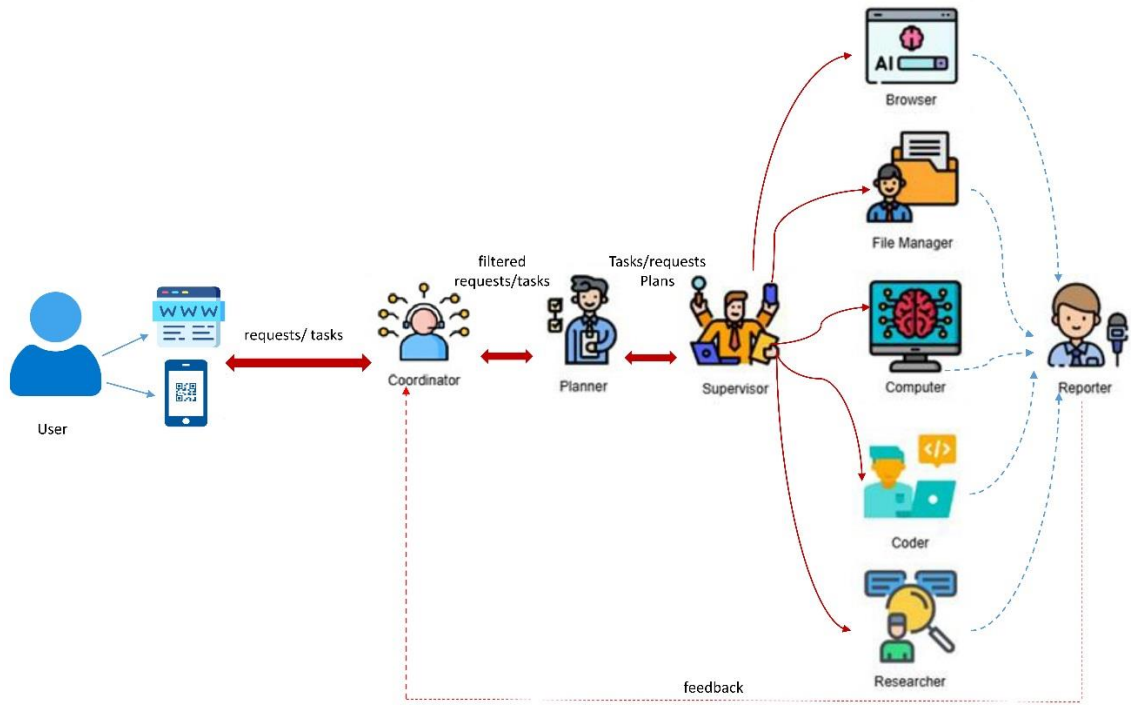

**Figure 1. Autonoma Multi-Agent Orchestration Framework.**

## 3.1 Coordinator agent

The Coordinator Agent serves as the conversational entry point and gatekeeper of the system. Built on GPT-4o, its primary role is to manage initial interactions by handling greetings and small talk in a professional and approachable manner, while ensuring safety through the polite rejection of harmful or inappropriate requests. When user input is unclear, the agent seeks clarification to gather sufficient context before proceeding. For all queries beyond its scope, the Coordinator seamlessly delegates by issuing a direct *handoff_to_planner()* command, transferring responsibility to the specialized planning module. In this way, the Coordinator maintains a balance between user engagement, safety, and efficient task routing, ensuring smooth communication across the multi-agent framework.

## 3.2 Planner agent

The Planner Agent, powered by GPT-4o and Deepseek R1, is responsible for transforming user requirements into structured, executable workflows. It begins by restating the user's request as a thought, then decomposes complex queries into actionable steps. These tasks are passed to the Supervisor Agent, which delegates them to the appropriate specialized agents. By coordinating this process, the Planner serves as the central orchestrator of the system, bridging user intent with precise, multi-agent task completion.

**Table 4.** Summary of agent roles, models, and functions within the Autonoma multi-agent system.

| Agent | Model | Responsibilities |
|---|---|---|
| **Coordinator** | GPT-4o | Routes queries, validates input |
| **Planner** | GPT-4o / DeepSeek R1 | Designs workflows, decomposes tasks |
| **Supervisor** | GPT-4o | Dispatches tasks, tracks execution |
| **Researcher** | GPT-4o | Web search, crawling, and data extraction |
| **Coder** | GPT-4o | Scripting, testing, and CLI operations |
| **Browser** | GPT-4o | Chrome automation and web interaction |
| **File Manager** | GPT-4o | File system operations (read, write, organize) |
| **Computer** | GPT-4o | Full desktop control with OCR integration |
| **Reporter** | GPT-4o | Summarizes results and generates structured reports |

## 3.3 Supervisor agent

The Supervisor agent, built on GPT-4o, manages and coordinates a team of specialized worker agents (Researcher, Coder, Browser, File Manager, Computer, and Reporter). For each workflow received from the Planner, it analyzes the tasks, selects the appropriate agent to execute them, and monitors progress. When multiple agents are involved, the Supervisor reviews outputs, ensures correct task handoffs, and either assigns the next step or marks the workflow as complete. This makes it the central manager of execution, ensuring accuracy, continuity, and task completion across the multi-agent system.

### 3.4 Browser agent

The Browser Agent, built on GPT-4o, is a web-interaction specialist that translates natural-language instructions into explicit, step-by-step browser actions. It can navigate to URLs, click links and buttons, type into forms, scroll pages, and extract targeted information, always returning the sequence of actions in clear natural language. Responses mirror the user's language to maintain consistency and usability across workflows.

### 3.5 File Manger agent

The File Manager Agent, built on GPT-4o, is designed to handle Windows file system operations with precision and safety. It can copy, move, delete, search, list directories, and read or write files by translating user requests into structured file operations. The agent plans and executes tasks step by step, confirms before performing destructive actions, and provides clear operation summaries, ensuring secure and efficient local file management.

### 3.6 Coder agent

The Coding Agent, powered by GPT-4o, functions as a software engineer proficient in Python and Bash within a Windows environment. It analyzes user requirements, writes and executes scripts, tests for edge cases, and documents its methodology. Operating through *PowerShell* with access to libraries such as *pandas*, *numpy*, *scipy*, *matplotlib*, *scikit-learn*, and *statsmodels,* as well as *yfinance* for market data, the agent handles data processing, calculations, and analysis tasks. By combining coding, execution, and result presentation, it streamlines software development and computational workflows.

## 3.7 Researcher agent

The Research Agent, powered by GPT-4o, is designed to analyze problem statements, plan research strategies, and execute them using a wide range of specialized tools. It employs resources such as *Tavily, Crawl*, and *DuckDuckGo* for web searches, *Wikipedia* and *YouTube* for domain-specific knowledge, *Yahoo Finance News* for financial insights, and *OpenWeatherMap* for environmental data. With the *HyperBrowser Extract* tool, it can capture and structure information directly from web sources. Its responsibilities include gathering external data, crawling, scraping, and returning structured, context-rich findings in the user's preferred language. By combining problem understanding with targeted tool use, the Research Agent delivers accurate, timely, and actionable insights across multiple domains.

## 3.8 Computer agent

The Computer Agent, powered by GPT-4o, is a Windows operations specialist capable of controlling the desktop environment and managing complex system interactions. It can open applications, navigate the interface, simulate mouse and keyboard actions, and handle file or web navigation. Additionally, it captures and processes screenshots, enabling seamless automation of tasks across the computer's graphical interface.

## 3.9 Reporter agent

The Reporting Agent, powered by GPT-4o, functions as a professional reporter that transforms completed workflows into clear, structured, and human-readable outputs. It compiles information into executive summaries, key findings, detailed analyses, and conclusions with actionable recommendations. By citing sources and distinguishing between factual evidence and analytical interpretation, it ensures accuracy and transparency. Its core responsibility is to aggregate outputs from all agents into a comprehensive final report, while also documenting failed tasks in log files and providing recommendations for resolving errors. This makes it the system's primary channel for delivering polished, reliable, and decision-ready insights.

## 4. Autonoma modularity design

The Autonoma system is designed with a modular, layered architecture that promotes scalability, flexibility, and maintainability. At the client layer, users can interact through a responsive web interface built with *React, Next.js, and TailwindCSS* or via a mobile application accessed through QR code scanning. The API layer, implemented in *FastAPI,* manages chat streaming, conversation organization, and persistence. Above this, the agent orchestration layer (i.e. powered by *LangGraph*) coordinates nine specialized agents, each handling distinct tasks such as browsing, coding, research, reporting, and file or desktop management. Finally, the storage layer ensures persistence by organizing chat histories, browser screenshots, and media files in a structured *JSON*-based system. The modularity of the design allows each layer to function independently while integrating seamlessly, a structure that supports efficient execution of complex workflows and provides a foundation for future extensibility (see Figure 2).

## 5. Autonoma workflow process

The proposed multi-agent Autonoma workflow operates through a structured pipeline designed to balance efficiency, reliability, and transparency (see Figure 3). At the beginning, the system receives user prompts, which are parsed by the Coordinator to extract intent, scope, and contextual cues such as language or urgency. This lightweight classification ensures casual conversations are distinguished from actionable tasks, thereby preventing unnecessary computational overhead.

Once the intent is clarified, the Coordinator routes the input to the appropriate component, i.e. either a casual-chat handler or the Planner, based on decision rules that leverage keyword triggers, entity recognition, and user history. Genuine task requests are then decomposed by the Planner into an optimized sequence of subtasks. This planning stage may incorporate preliminary information gathering through external search tools and applies graph-based strategies to maximize both efficiency and parallelism.

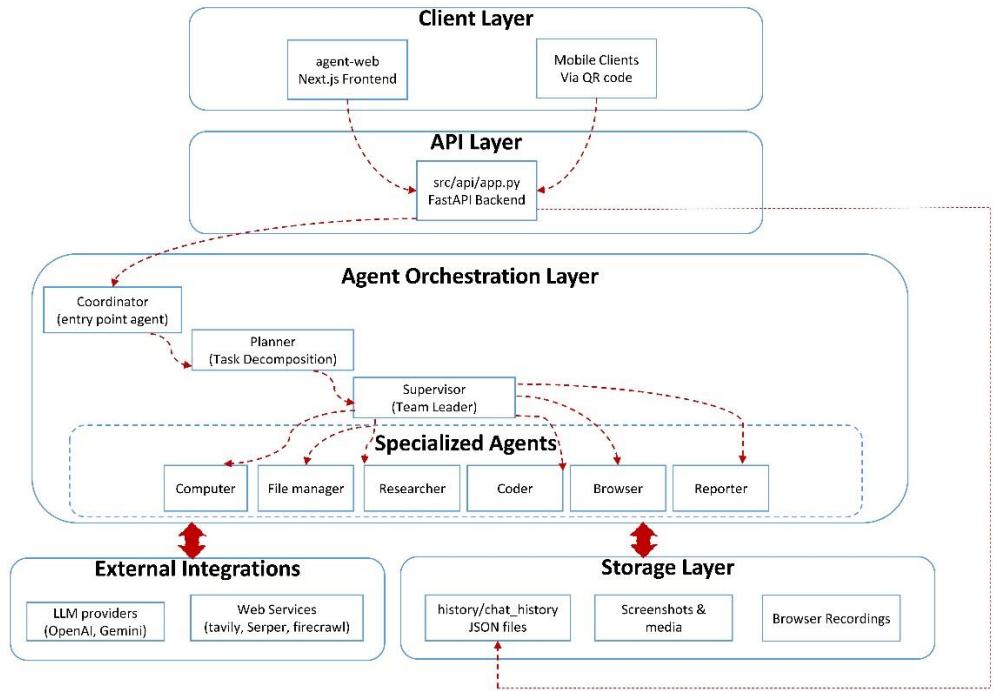

**Figure 2. Modular architecture of the Autonoma system.**

With a plan established, the Supervisor assumes control, allocating subtasks to specialized worker agents such as Researchers, Coders, or Browsers. Task progress is monitored continuously through acknowledgments and periodic health checks, while built-in retry mechanisms and error-recovery strategies ensure resilience against transient failures. Execution proceeds either sequentially or in parallel depending on task dependencies, with adaptive load-balancing strategies employed to optimize resource use and maintain scalability.

As tasks complete, the Reporter aggregates the outcomes into a structured and reproducible summary. This integrated report highlights key findings, generated artifacts, and any anomalies detected during execution. Finally, results are relayed back to the user via live feedback channels, enabling transparency, user oversight, and real-time intervention when necessary. This feedback loop not only enhances trust but also supports iterative refinement of workflows, ensuring both robustness and adaptability of the system in practical deployment.

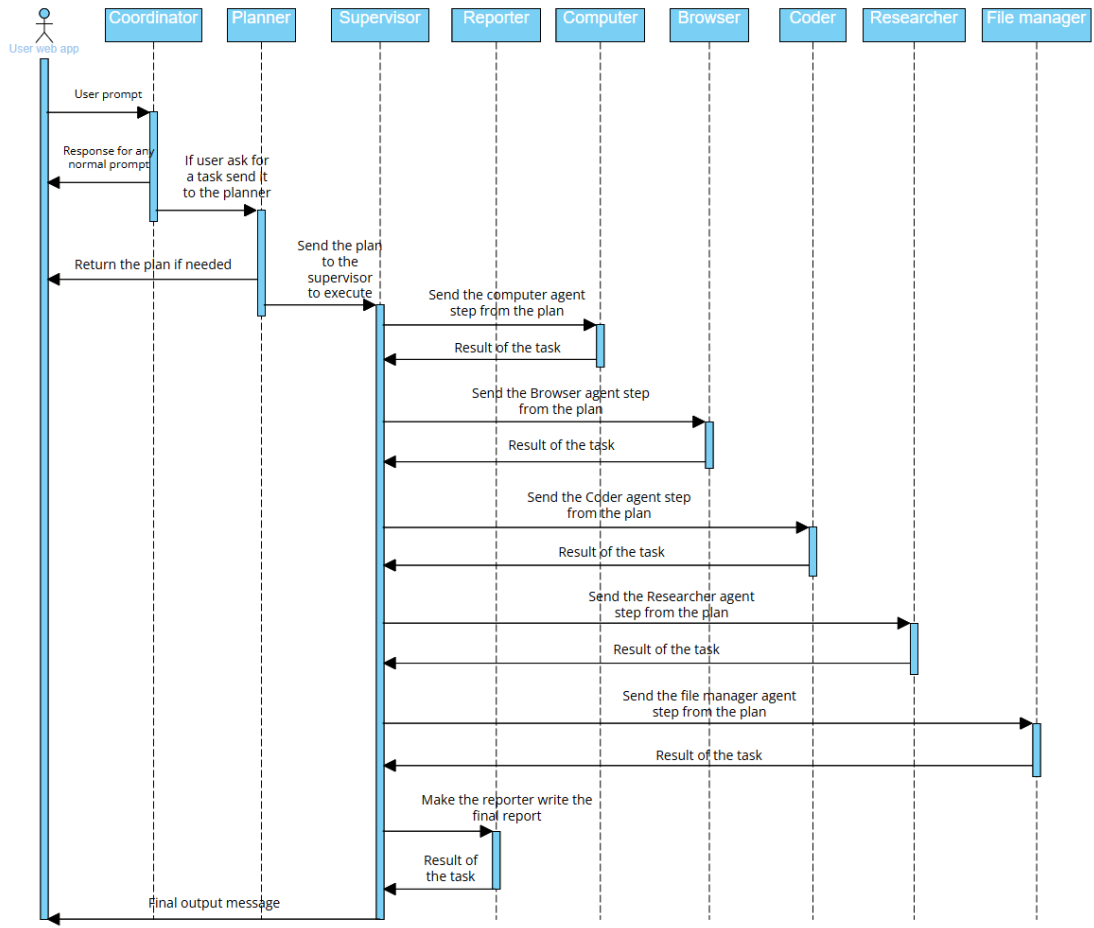

**Figure 3. Autonoma System Workflow.**

# 6. Autonoma Technology Stack, Security and Extensibility Considerations

## 6.1 Technology Stack

The proposed multi-agent Autonoma assistant system is developed based on a modular and layered technology stack designed to ensure performance, scalability, and developer productivity. At the frontend, *Next.js* is employed for server-side rendering and routing, combined with *React* for component-based user interface development. Styling is facilitated through *Tailwind* CSS, which enables responsive design with minimal overhead. Backend orchestration is achieved through *LangChain* and *LangGraph* [30], which support dynamic agent workflows by chaining Large Language Model (LLM) calls and managing graph-based state dependencies, respectively. Task planning and summarization are powered by GPT-4o, chosen for its low-latency multi-modal reasoning, while Deepseek R1[31] is integrated for deep inference and complex workflow optimization. For perception tasks, the computer vision agent utilizes *YOLOv8* [32], a high-performance, real-time object detection framework. Communication between system components is enabled through *HTTP POST* endpoints for prompt submission and *WebSockets* for event-driven bidirectional updates, with optional voice input captured via the Web Speech API to support real-time transcription.

## 6.2 Security

Security and privacy are central to the design of the Autonoma orchestration framework. To minimize exposure to external threats, the system enforces LAN-only isolation using IP filtering and QR-code authentication. Each agent operates within a sandboxed containerized environment with restricted privileges, reducing the risk of escalation or cross-agent interference. Comprehensive audit trails are maintained, recording timestamps, agent actions, input/output artifacts, and GUI interactions, thereby ensuring accountability and facilitating regulatory compliance. Moreover, the system enforces a least-privilege access model, restricting agents and services to only the permissions required for operation. These measures collectively ensure resilience, traceability, and adherence to best practices in secure distributed AI systems.

## 6.3 Extensibility

Extensibility is a core design principle of the Autonoma framework, allowing it to adapt to evolving requirements and third-party integrations. The architecture supports modular agent registration through a plugin-based mechanism, enabling new agents to be defined and deployed at runtime without modifying the core system. Developers can inject custom hooks and interceptors at various workflow stages to introduce domain-specific logic such as data validation, transformation, or notifications, ensuring flexibility in diverse application contexts. Additionally, the system is fully containerized using *Docker*, enabling seamless deployment across on-premise and cloud infrastructures. When deployed on orchestration platforms such as *Kubernetes*, the system benefits from horizontal scalability, rolling updates, and environment consistency features that align with modern *MLOps* practices.

## 7. Experimental Results and Future Challenges

### 7.1 Experimental Results

To evaluate the performance and robustness of the proposed multi-agent Autonoma assistant system, we conducted a comprehensive assessment across multiple dimensions, including usability, reliability, responsiveness, agent collaboration, multilingual support, and security (see Table 5).

**Table 5. Evaluation Metrics and Results of Autonoma Multi-Agent Assistant System.**

| Metric | Description | Result |
|---|---|---|
| **Task Completion Rate** | Percentage of tasks successfully completed without errors. | 97% success rate across 500 test cases |
| **Frontend Latency** | Average time from user input to visible system response. | 1–2 seconds, depending on network conditions |
| **Agent Coordination Efficiency** | Effectiveness of task decomposition and delegation among agents. | 98% successful inter-agent task handoffs |
| **Multilingual Interaction** | Ability to switch seamlessly between Arabic and English. | 100% success rate without page reloads |
| **Security Validation** | Effectiveness of LAN isolation and sandboxing against unauthorized access. | 0 breaches detected in penetration tests |

The evaluation of the proposed Autonoma multi-agent AI assistant system demonstrates its effectiveness across several critical performance dimensions, confirming both practical usability and technical robustness. From a usability standpoint, the system enabled users with limited technical expertise to complete complex workflows with minimal training. This suggests that the interface design, combined with integrated guidance mechanisms, successfully facilitates intuitive interaction, a significant finding, given that usability remains a common barrier to the adoption of sophisticated AI systems. In terms of reliability, Autonoma achieved a 97% task completion rate across a diverse set of 500 test cases. This high degree of operational robustness can be attributed to the supervisory orchestration layer, which implements systematic error detection, retry policies, and fault-tolerance mechanisms. This performance is competitive when compared with existing multi-agent orchestration frameworks reported in the literature.

System responsiveness was measured through frontend latency, which averaged between 1–2 seconds under typical network conditions. Although this is slightly above the ideal sub-second threshold, it remains acceptable for most practical applications. Future optimizations, such as lightweight model distillation or client-side caching, could further reduce this latency.

The framework also exhibited strong collaborative performance, with a 98% success rate in inter-agent task handoffs. This reflects the efficacy of the underlying LangChain and LangGraph-based orchestration engine in ensuring coherent task decomposition, delegation, and execution. Such reliable coordination is essential for scalability, particularly in environments involving concurrent workflows or multi-step tasks.

Multilingual capability (i.e. specifically, seamless switching between Arabic and English) was validated with a 100% success rate without requiring page reloads. This demonstrates the robustness of the localization layer and significantly enhances the system's accessibility and cross-regional applicability.

Security was assessed via penetration testing, which confirmed the effectiveness of the defense-in-depth strategy, including LAN-only operation and sandboxed agent execution. No breaches were detected, underscoring the architectural strength in protecting data and workflow integrity. While resilience against advanced persistent threats requires further investigation, the current security baseline is solid. Collectively, these results validate Autonoma as a reliable, secure, and user-friendly framework for multi-agent AI orchestration.

## 7.2 Future Challenges

While the Autonoma framework provides a robust foundation for multi-agent AI systems, its capabilities can be substantially advanced by addressing persistent research challenges in scalability, intelligence, and safety. To handle environments of increasing complexity and scale, future work will focus on advancing agent collaboration paradigms. This includes investigating decentralized coordination mechanisms, dynamic role assignment, and formal conflict-resolution protocols. Such advancements are crucial for enabling agents to make independent, yet coherent, decisions in real-time, thereby improving the system's overall resilience and efficiency in large-scale, dynamic task environments.

A key direction is the infusion of adaptive learning capabilities to allow the system to personalize its interactions and improve over time. By integrating continuous learning models that leverage user feedback and interaction histories, agent behavior can be refined to produce more tailored and effective outcomes. This aligns with the long-term goal of creating AI assistants that evolve with user preferences and operational contexts. Expanding the modalities of interaction is critical for more natural and immersive user experiences. Future research will explore the integration of capabilities such as gesture recognition, facial expression analysis, and haptic feedback. Realizing this vision will require concurrent advances in multi-sensor fusion and low-latency data processing pipelines.

Supporting larger workloads and user bases necessitates a shift towards distributed, cloud-native architectures. This must be coupled with the development of optimized algorithms for parallel processing and dynamic resource orchestration. As the system's scope expands, so must its security posture. Future work will integrate advanced cryptographic techniques, intrusion detection systems, and privacy-preserving methodologies like federated learning to ensure data integrity and user trust.

Tailoring Autonoma for specialized domains such as healthcare, education, and finance will require the development of domain-specific agents and robust interoperability protocols. Adoption of open standards, such as the Model Context Protocol (MCP), will be instrumental for seamless and secure integration with external tools and data sources. To increase operational robustness, we plan to embed mechanisms for learning from failures. Techniques from reinforcement learning could allow agents to diagnose root causes of errors and iteratively refine their action plans, thereby reducing error recurrence and enhancing autonomous problem-solving. The system's versatility can be systematically extended by developing a broader portfolio of specialized worker agents. Potential areas for development include advanced data analysis, content generation, and real-time system monitoring, enabling the automation of a wider array of complex, cross-functional tasks. Finally, the pervasive deployment of multi-agent systems demands rigorous attention to their ethical and societal implications. A dedicated research thread will address critical issues including informed user consent, system transparency, algorithmic fairness, and the mitigation of unintended consequences. This work is essential to ensure that the system's evolution remains aligned with human values and contributes positively to society.

## 8. Conclusions

This paper introduced Autonoma, a multi-agent AI assistant system designed to bridge the critical gap between conversational AI that proposes actions and a functional system that executes them within a user's native environment. Through its hierarchical agent architecture, comprising a Coordinator, Planner, Supervisor, and a suite of specialized Worker agents, Autonoma demonstrates a robust methodology for translating natural language prompts into structured, executable workflows. Our evaluation validates the framework's practical utility and technical robustness. The system achieved a high task completion rate of 97%, demonstrated efficient agent collaboration with a 98% successful handoff rate, and proved accessible to non-technical users through its intuitive chat interface. Furthermore, the implementation of a secure, LAN-confined environment with defense-in-depth measures establishes a solid foundation for safeguarding data and workflows. These results collectively confirm that Autonoma provides a reliable, user-friendly, and scalable platform for multi-agent orchestration. While these results are promising, achieving fully autonomous, large-scale agent systems requires further progress in latency, adversarial resilience, and adaptive learning. The Autonoma framework provides a foundational and extensible platform to address these critical challenges, representing a significant advance toward practical and pervasive human-AI collaboration.

## Statements and Declarations

### Data availability

To facilitate result reproducibility, the system's source code is available on GitHub at https://github.com/eslam-reda-div/Autonoma, and a comprehensive wiki description can be found at: https://deepwiki.com/eslam-reda-div/Autonoma.

### Competing Interests

The authors declare no competing interests.

### Authors' contributions

ER, MY, & SE: Conceptualization, Methodology, Software, Validation, Formal analysis, Investigation, Resources, Writing – original draft, Writing – review & editing. SE: Supervision, Project administration.